\documentclass{article}

\usepackage[final]{neurips_2022}

\usepackage[utf8]{inputenc} % allow utf-8 input
\usepackage[T1]{fontenc}    % use 8-bit T1 fonts
\usepackage{hyperref}       % hyperlinks
\usepackage{url}
\usepackage{booktabs}       % professional-quality tables
\usepackage{amsfonts}       % blackboard math symbols
\usepackage{nicefrac}       % compact symbols for 1/2, etc.
\usepackage{microtype}      % microtypography
\usepackage{xcolor}         % colors
\usepackage{subcaption}
\usepackage{graphicx}
\usepackage{tablefootnote}

\usepackage{fourier} 
\usepackage{array}
\usepackage{makecell}
\title{Large scale traffic forecasting with gradient boosting \\
    \large Traffic4cast 2022 challenge \\
    \large Team Bolt
}

\author{%
  Martin Lumiste \\
  Bolt Technology\\
  Tallinn, Estonia\\
  \texttt{martin.lumiste@bolt.eu} \\
  \And
  Andrei Ilie\thanks{cilie@fmi.unibuc.ro} \\
  Bolt Technology, University of Bucharest \\
  Bucharest, Romania \\
  \texttt{andrei.ilie@bolt.eu}
}

\begin{document}

\maketitle

\begin{abstract}
    Accurate traffic forecasting is of the utmost importance for optimal travel planning and for efficient city mobility. IARAI \footnote{The Institute of Advanced Research in Artificial Intelligence} organizes Traffic4cast, a yearly traffic prediction competition based on real-life data \footnote{\url{https://www.iarai.ac.at/traffic4cast/}}, aiming to leverage artificial intelligence advances for producing accurate traffic estimates. We present our solution to the IARAI Traffic4cast 2022 competition, in which the goal is to develop algorithms for predicting road graph edge congestion classes and supersegment-level travel times. In contrast to the previous years, this year's competition focuses on modelling graph edge level behaviour, rather than more coarse aggregated grid-based traffic movies. Due to this, we leverage a method familiar from tabular data modelling - gradient-boosted decision tree ensembles. We reduce the dimensionality of the input data representing traffic counters with the help of the classic PCA method and feed it as input to a LightGBM model. This simple, fast, and scalable technique allowed us to win second place in the core competition. The source code and references to trained model files and submissions are available at \url{https://github.com/skandium/t4c22}.
\end{abstract}

\section{Introduction}

The Institute of Advanced Research in Artificial Intelligence proposes a yearly traffic forecasting challenge at NeurIPS, Traffic4cast. This competition has the goal of uncovering and deepening the application of artificial intelligence methods in the traffic forecasting domain. 

This year, the goal of the competition was to leverage public vehicle counter data \footnote{Counts of vehicles passing through map nodes, measured automatically with the aid of electronic devices.} for producing city-wide traffic estimates. There is wide applicability to this, as traffic counters are easily accessible sensors that many cities already use. Efficient and accurate methods to infer city-wide segment-level traffic patterns on top of live counter data would be highly useful for optimising city mobility and travel planning. These would also represent an affordable, easy-to-scale approach, as opposed to having traffic cameras in most city intersections or having live data streams from many probes covering the city continuously.

The core and extended Traffic4cast tracks featured somewhat different prediction problems. The core track contains a multi-class classification task at a segment level, where the goal is to predict one out of three predefined congestion classes (green, yellow, or red), which represent traffic intensities. The training labels are sparse relative to the entity we are predicting: in any given training temporal slice, we might observe labels for only about 5\% of the edges that we will need to make predictions for. The evaluation metric used is weighted cross entropy, placing more emphasis on the classes that appear more rarely, with the highest weight for red (congested segments), then for yellow, and then for green classes.

The extended track proposes a regression task at a supersegment \footnote{Supersegments are sequences of segments created according to some given methodology, connecting main intersections.} level, optimizing for mean absolute error. The training data for this task consists of dense labels (we observe all supersegment labels at all times).

We reduce both problems to a tabular format, capturing intrinsic features of the road entities we want to create predictions for (e.g: latitude, longitude, target encoding of historical speeds) and features representing the local and city-wide traffic states. Most interestingly, we capture the city-wide traffic state by performing dimensionality reduction on top of the available counter data at any temporal slice. This, together with the gradient-boosted modelling approach we used, constitute the key elements of our solution. The methodology remains similar for both tracks, but notes are provided below for where slight task-specific modifications were required.

\section{Solution}

\subsection{Motivation}

It is well known in the literature and applied machine learning competitions that gradient-boosted tree ensembles generally outperform neural networks on tabular data \citet{tabular21} \citet{tabular22} \citet{borisov21} \citet{lundberg20}, especially when one considers the accuracy and training time trade off. Recently, this has also been shown for large scale multivariate time series forecasting problems. \citet{makridakis22}
On the other hand, fields like computer vision and natural language processing are dominated by deep learning approaches due to their unstructured data which must first be projected into a more easily separable space by the first layers of the networks. It is then a natural question whether spatio-temporal graph problems such as traffic prediction are close enough to tabular that they can benefit from gradient boosting. Our main hypothesis was that this is the case.

Another core motivation for our solution was that traffic is highly seasonal. Therefore, features like the day of the week and the time of the day would be greatly informational. Unfortunately, such temporal features are not available at test time, so we tried to recover them implicitly from the city traffic state. Our hypothesis was that the global city traffic context would be beneficial for modelling as it not only acts as a proxy for temporal features, but is even more powerful due to being robust against distribution shifts that may occur between train and test sets.

\begin{figure}[h!]
\includegraphics[width=\textwidth]{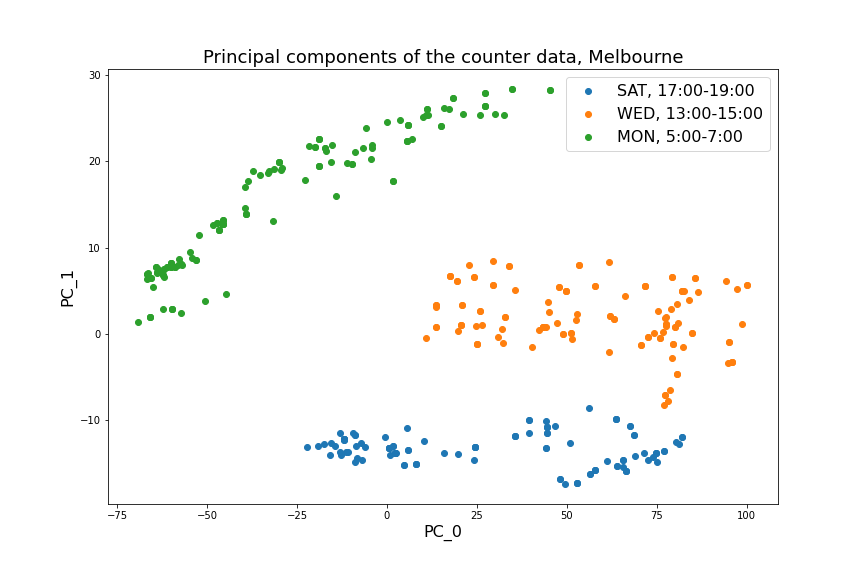}
\caption{The first two principal components of the last traffic counter measurements are highly separable for periods reflecting different traffic conditions.}
\label{fig:PCA_separable}
\end{figure}

% \begin{figure}[h!]
% \centering
% \begin{subfigure}{0.99\textwidth}
%     \includegraphics[width=\textwidth]{melbourne_diff_PC_0_PC_1.png}
%     \caption{First subfigure.}
%     \label{fig:first}
% \end{subfigure}
% % \hfill
% % \begin{subfigure}{0.49\textwidth}
% %     \includegraphics[width=\textwidth]{melbourne_lunchtime_wk_PC_1_PC_5.png}
% %     \caption{Third subfigure.}
% %     \label{fig:third}
% % \end{subfigure}
        
% \caption{Subreferences in \LaTeX.}
% \label{fig:figures}
% \end{figure}

\subsection{Data preprocessing}

The dataset provided by the competition consisted of three main sources: a) the vehicle counter time series with roughly four thousand counters per city at a granularity of 15 minutes, b) task-specific labels for road segments, also in 15-minute intervals (shifted one step ahead of the counter observation) and c) static road graph attribute metadata (e.g the coordinates of nodes or the road types of edges). Aggregated traffic movies similar to previous years are also provided, but we do not use these in our solution.

As the counter volumes are provided as a time series, the first step is to transform those into numeric features. Traffic4cast 2021 showed that using only the most recent 1 hour of data carries enough signal (\cite{t4c21}). Due to this, we calculate only the sum of the past hour and the very last measurement as numeric features. To reduce the relatively high dimensions of the counter data (Melbourne has 3980 counters and 10260 measurement intervals) and to embed the global city \textbf{traffic context}, we compute the principal components of the \(k \times t\) counter matrix $\mathbf{V}$ where $k$ is the number of counters in the city and $t$ the number of observations. We leverage the \(k \times k\) covariance matrix, meaning our resulting principal components are time series. In practice, we use the first 8 principal components of the last volume and the first 5 principal components of the sums - confirmed on the  validation set to be the optimal strategy. We observe that the first few principal components explain well the data variance. For example, in Melbourne's case, the top 5 principal components of the last hour's sum of counters features explain 93.12\% of the variance.

For embedding the \textbf{spatial context} of edges and supersegments, we mostly use raw $x$ and $y$ coordinates. For edges, we use the start node as its representative point. However, as supersegments are defined as a list of nodes which can be multiple kilometres long, we use its 2-dimensional medoid as the representative point. Later experiments showed that also using the start and end points of supersegments together with its medoid and distance can be beneficial. This is because the supersegments start and end by definition at popular intersections. As the set of these intersections has low cardinality (roughly 500), there is value in trying to learn to profile them.

On the other hand, the cardinality of the modelled entity itself is very large (roughly $100.000$ segments in core, thousands of supersegments in extended). We save model training bandwidth by precomputing segment characteristic \textbf{target encoding} features based on training set labels. Moreover, we calculate segment characteristic features conditional on the traffic state of the whole city. This is equivalent to the IARAI baseline provided in the extended track - overall ETAs per supersegment, conditional on city traffic cluster - but used as a feature, rather than prediction. As decision tree ensembles are high-capacity learners, we additionally guard against the risk of data leakage by only calculating target encoded features based on the train set if an edge had a significant enough amount of observations within that cluster bucket, otherwise overwriting the values with fallback constants. We also leverage the raw edge-level speed labels (data source for both core and extended track labels) and apply this methodology to calculate the median speed and free flow speed per edge.

Finally, to embed the graph nature of the problem, we implement a simple spatial weighting method inspired by graph \textbf{message passing}. Namely, given the counter matrix $\mathbf{V}$ from above, let us define a symmetric weight matrix \( \mathbf{B} \in \mathbb{R}^{k \times k} \) where each row element is a normalized score summing up to one. Each element $b_{i, j}$ gives the spatial relevance of counter $i$ with respect to $j$. To find $\mathbf{B}$, one could apply any monotonically decreasing function in distance row-wise on the pairwise distance matrix of counters. For example, we experiment with applying softmax on the inverse distances. By multiplying \( \mathbf{W} = \mathbf{V}^T \mathbf{B} \in  \mathbb{R}^{t \times k} \) we have that \( w_{t, i} = \sum_{j}^{K} v_{t, j} \times b_{i, j} \) so each row gives the spatially weighted context around each counter. Note that this is a fairly generic approach, as $\mathbf{B}$ could be based on any distance metric such as Euclidean or graph hop distances between the counters. By making $\mathbf{B}$ a sparse matrix, we could enforce nearest neighbour features. We experiment with multiple Euclidean and nearest neighbour weighting methods and join the resulting matrixes $\mathbf{W}$ to our dataset based on the nearest counter to the road. 

We give a high level overview of the data preprocessing we do in Appendix \ref{appendix:data_preproc}, Figure \ref{fig:circuit}.
\subsection{Training}

Gradient boosting can be thought of as performing gradient descent in functional space \cite{friedman99} by $\hat{y}_i = \sum_{m=1}^{n} h_m (x_i) + h_0$ where the weak estimators $h_m$ use the residuals of the previous ensemble prediction $h_{m-1}$ as labels and $h_0$ is usually a constant. But this is not a requirement, we can use any predictor as $h_0$ to provide a robust basis for the model to improve upon. To speed up model convergence, we initialize it from a smart baseline - the target encoding features described above. In practice, this means that for the core track, our first iteration consists of the weighted logits coming from the train set average class distributions. For the extended part, we use traffic conditional median ETAs, similar to the IARAI baseline. This helps the model become not only more accurate than the baseline already in a few iterations, but also converge to a better end result, likely due to the avoided complexity that making splits for each road entity would entail.

We use the highly efficient gradient boosting library LightGBM \citet{lgb} for training. Hyperparameters are tuned once according to a predefined stepwise Optuna strategy \citet{ozeki20}. We only tune the hyperparameters on a small subset of extended track data for Melbourne and use these as a sane baseline for every model. Thereafter, the only hyperparameters that we choose are the number of leaves in individual decision trees ($num\_leaves$) and the number of gradient boosting iterations ($num\_iters$). Surprisingly, for the core track, we found that the $num\_leaves$ parameter which controls the tree complexity by the number of unique terminal leaf nodes could be increased almost arbitrarily, speeding up convergence significantly and not hurting the holdout test set accuracy. We therefore use 5k leaves for Melbourne and 10k for Madrid and London.  

We simulate interleaved validation week selection, keeping in line with how test set is separated. For example, Melbourne has the following weeks of the year available in the training data: $[23, 25, 27, 29, 31, 33, 35, 37, 39, 41, 43, 45, 47, 49, 51, 53]$.
We use weeks $[25, 33, 41, 49]$ for validation, and the rest for training.

We generally prototyped all city models locally on a random subsample of the data before scaling them to a cloud instance. Table \ref{train-summary} shows the summary of our training strategy \footnote{Extended models were trained on a laptop with swap enabled, actual memory usage is higher}. We think that the quick training times with no GPU requirements are what make LightGBM an attractive solution to traffic forecasting, lowering the technical and financial barriers of entry in this field. For even larger problems, there exists a distributed implementation of LightGBM which has been shown to scale up to a 1.7 billion row dataset \citet{lgbdist}.

\begin{table}[h]
  \caption{Results}
  \label{table:results}
  \centering
  \begin{tabular}{cccccccc}
    \toprule
 \textbf{Track} & \textbf{Model} & \textbf{PCA}    & \textbf{init\_score}    & \textbf{Target encoding} & \textbf{Tuned} & \textbf{Score} \\
 \hline
 \hline
  \textit{Core}     &     &               &             &                &           & \makecell[c]{Weighted \\cross-entropy}\\
 \hline
     & IARAI baseline \tablefootnote{Vanilla GNN}   &               &             &                &           & 0.8978\\
     & Bolt              &  X            &             &                &          & 0.8877 \\
     & Bolt              & X             & X            &                &          & 0.8692 \\
     & Bolt              & X             & X            & X               & X         & \textbf{0.8497}  \\
\hline
 \textit{Extended}     &     &               &             &                &           & \makecell[c]{Mean \\ absolute error}\\
 \hline
 & IARAI baseline \tablefootnote{Average ETA per supersegment, conditional on city traffic cluster}    &               &             &                 &          & 67.21  \\
 & Bolt              &               & X            &                 &          & 65.03   \\
 & Bolt              &  X             & X            &                 &          & 63.56   \\
 & Bolt              &  X             & X            & X                &          & 61.58   \\
 & Bolt              &  X             & X            & X                & X         & \textbf{61.25}          
    % \bottomrule
  \end{tabular}
\end{table}
\subsection{Results}
Our approach ranked second in the core track and fourth in the extended one. The results are presented in Table \ref{table:results}, where we also introduce the ablation study of our models. We perform ablation on four of the key components of our solutions, which are shared across both core and extended models: using city-wide PCA features (\textbf{PCA}), feeding (traffic conditional) target encodings as initialization score to LightGBM (\textbf{init\_score}), using (traffic conditional) target encodings (at multiple granularities) as features to the model (\textbf{Target encoding}), and performing hyperparameter tuning using Optuna and early stopping (\textbf{Tuned}).

\section{Discussion}

The biggest surprise to the authors was that message passing did not work at all. We found no meaningful improvement from adding locally weighted counter features, especially when compared to the city global principal component features. We experimented with different weighting methods and the most successful ones tended to be ones that still used the entire city's counter data and applied only a marginal penalty for high distances, essentially converging to city global features. We think that this can be partly explained by the heuristic nature of our feature engineering, while a weighting schema learned by a GNN should work better. However, it should not be discounted that decision trees are natural interaction learners - indeed, the original motivation for the tree method came from analyzing interactions in survey data \citet{og_tree}. It seems that passing a combination of city global context features and the spatial coordinates of roads is enough for learning their interaction and therefore also the local context, conditional on the overall traffic state.  

Traffic is a spatio-temporal process and hiding the temporal data by competition design definitely limits any solution's potential accuracy. However, we find that our reduced form city context features are enough to recover the temporal features and act as powerful proxies for them. In a real-life setting, including both the time features and counter data should give the best result.

Unsurprisingly, target encodings were a critical component of making the models competitive. However, we additionally found that the degree to which one can leverage this signal is proportional to the density of the labels used. Due to this, we managed to develop very intricate target encoded features in the extended track, calculating multiple granularities of traffic quantiles for observations and using all of them as features, without any visible lack of generalization. On the core track, we are conditioning the target encoded features only on a binary "low" or "high" traffic regime, experiments with more granular partitions started to suffer due to leakage.

Although we found gradient-boosted decision tree ensembles to be competitive for this competition, it is worth pointing out some of this architecture's shortcomings as well. Firstly, there is no natural embedding method for decision trees. One could pass entity tokens as categorical features and LightGBM would try to apply Fisher grouping \citet{fisher58}, but this is essentially useless for a high cardinality feature such as $edge\_id$. In practice, we circumvent this by calculating target encoded features for each token we think is relevant. However, this is suboptimal, as we need to explicitly handle tokens with few values and there is an increased risk of data leakage because of using the labels directly. Closely related to the first problem is that it is difficult to embed sequential features with our tabular method. In the extended track, explicit node sequences provide an obvious sequential signal for predicting ETAs. An architecture such as an RNN would elegantly aggregate the sequential node embeddings into vector space. Therefore a neural network with a sequential input component should be competitive in the extended track, but an alternative hybrid approach, as shown by \citet{leontjeva16} would be to extract the sequential activations and use them as features in our tabular model. We did not experiment with these approaches in the competition and these could remain as future avenues of traffic research.

\newpage

%%%%%%%%%%%%%%%%%%%%%%%%%%%%%%%%%%%%%%%%%%%%%%%%%%%%%%%%%%%%
\bibliography{sources}

\appendix

\section{Data preprocessing overview}
\label{appendix:data_preproc}
\begin{figure}[h]
\caption{Summary of the preprocessed data used for training. We are using three main sources of data: counter data, training set labels and static map data. }
\label{fig:circuit}
\includegraphics[width=0.99\linewidth]{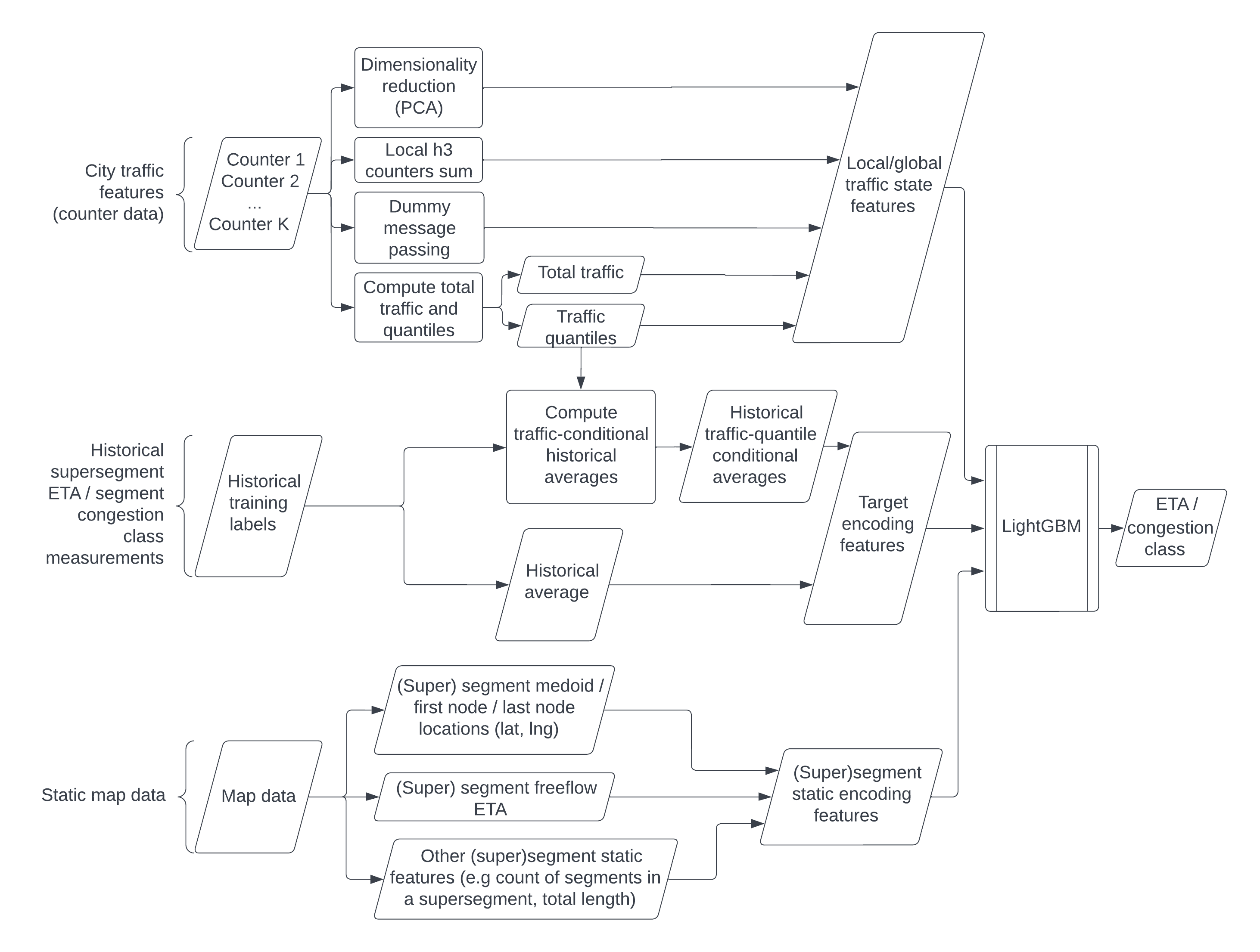}
\end{figure}

\section{Training summary}

Table \ref{train-summary} gives the LightGBM training summary.

\begin{table}[h]
  \caption{Training summary}
  \label{train-summary}
  \centering
  \begin{tabular}{llllllll}
    \toprule
\textbf{Track}    & \textbf{City}      & \textbf{Rows (Mil)} & \textbf{Features} & \textbf{RAM (GB)} & \textbf{Leaves} & \textbf{Iters} & \textbf{Time (h)} \\
Core     & Melbourne & 98              & 44            & 256               & 5k          & 250        & <1           \\
Core     & London    & 337             & 44            & 512               & 10k         & 450        & 4           \\
Core     & Madrid    & 482             & 44            & 768               & 10k         & 600        & 5           \\
Extended & Melbourne & 33              & 31            & 32                & 350         & 3200       & 1.5            \\
Extended & London    &  42             & 31            & 32                & 400         & 5700       & 2          \\
Extended & Madrid    &  41             & 31            & 32                & 350         & 4900       & 2           
    % \bottomrule
  \end{tabular}
\end{table}

\section{Feature importance}
\label{appendix:feature_imp}
\begin{figure}[h]
\caption{Exact Shapley decision tree feature importances - core, Melbourne \citet{lundberg20}}
\label{fig:shap}
\includegraphics[scale=0.4]{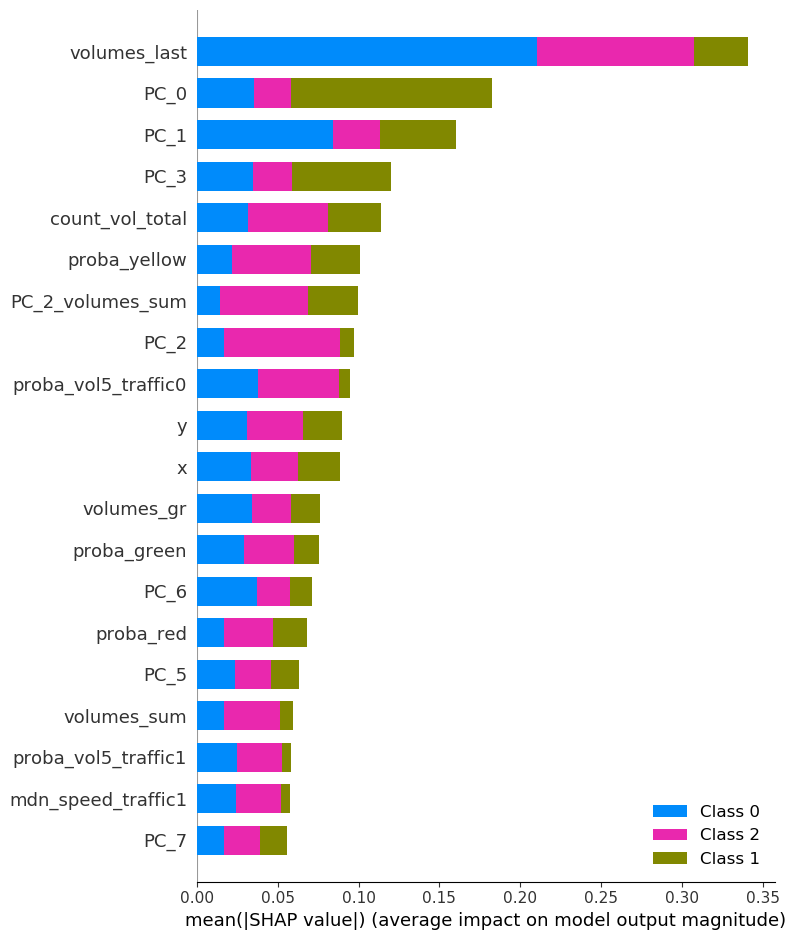}
\end{figure}

Figure \ref{fig:shap} visualizes the most important features (out of 44) of a fully trained model for core track, including their contributions to each congestion class. We see how city global features such as PCs and $volumes$ (simple city wide time based medians) dominate the predictions, with the spatial context in $x$/$y$ close behind. Everything else comes from edge specific target encodings. Edge attributes do not make it into the most relevant features.
 
\end{document}